\documentclass[lettersize,journal]{IEEEtran}
\usepackage{amsmath,amsfonts}
\usepackage{algorithmic}
\usepackage{algorithm}
\usepackage{array}
\usepackage[caption=false,font=normalsize,labelfont=sf,textfont=sf]{subfig}
\usepackage{textcomp}
\usepackage{stfloats}
\usepackage{url}
\usepackage{verbatim}
\usepackage{graphicx}
\usepackage{cite}
\usepackage{amsmath}
\usepackage{amsfonts}
\usepackage{float}
\usepackage{graphicx}
\usepackage{booktabs}
\usepackage{multirow}
\usepackage{epstopdf}
\usepackage{enumitem}
\usepackage{colortbl}
\usepackage[colorlinks=true, linkcolor=black, urlcolor=gray, citecolor=black]{hyperref}
\usepackage[table]{xcolor}
\usepackage{colortbl}
\definecolor{lightbluegray}{RGB}{235,245,255}
\definecolor{lightgreengray}{RGB}{235,245,235}
\usepackage{pifont}
\usepackage[utf8]{inputenc} 
\usepackage{mdframed} 
\usepackage{enumitem} 

\begin{document}

\title{WHERE and WHICH: Iterative Debate for Biomedical Synthetic Data Augmentation}


\author{Zhengyi Zhao$^1$, Shubo Zhang$^2$, Bin Liang$^1$, Binyang Li$^2$, Kam-Fai Wong$^1$\\$^1$The Chinese University of Hong Kong\\$^2$University of International Relations\\\texttt{\{zyzhao, kfwong\}@se.cuhk.edu.hk}}



\maketitle

\begin{abstract}

In Biomedical Natural Language Processing (BioNLP) tasks, such as Relation Extraction, Named Entity Recognition, and Text Classification, the scarcity of high-quality data remains a significant challenge. This limitation poisons large language models to correctly understand relationships between biological entities, such as molecules and diseases, or drug interactions, and further results in potential misinterpretation of biomedical documents. To address this issue, current approaches generally adopt the Synthetic Data Augmentation method which involves similarity computation followed by word replacement, but counterfactual data are usually generated. As a result, these methods disrupt meaningful word sets or produce sentences with meanings that deviate substantially from the original context, rendering them ineffective in improving model performance. To this end, this paper proposes a biomedical-dedicated rationale-based synthetic data augmentation method. Beyond the naive lexicon similarity, specific bio-relation similarity is measured to hold the augmented instance having a strong correlation with bio-relation instead of simply increasing the diversity of augmented data. Moreover, a multi-agents-involved reflection mechanism helps the model iteratively distinguish different usage of similar entities to escape falling into the mis-replace trap. We evaluate our method on the BLURB and BigBIO benchmark, which includes 9 common datasets spanning four major BioNLP tasks. Our experimental results demonstrate consistent performance improvements across all tasks, highlighting the effectiveness of our approach in addressing the challenges associated with data scarcity and enhancing the overall performance of biomedical NLP models.
\end{abstract}


\section{Introduction}
Biomedical Natural Language Processing (BioNLP) is essential for advancing medical research and improving healthcare by extracting valuable information about relationships between biological entities like genes, diseases, and drugs. It generally requires a high degree of accuracy due to the critical nature of the biomedical field. The extracted information will often be used in medical research, diagnostics, and treatment decisions, and directly influence the performance. Even minor inaccuracies or imprecise statements can lead to severe consequences, such as misinterpretation of drug interactions, incorrect disease classifications, or flawed research conclusions.

Therefore, in BioNLP tasks, the scarcity of high-quality annotated data poses a significant challenge. However, nowadays, current approaches employ large language models (LLMs) to capture the intricate relationships within the biomedical domain, which require a substantial amount of precise data \cite{ding2020daga, liu2021mulda, zhou2021melm}. And insufficient or low-quality data will lead to significant misinterpretations, such as incorrect associations between drugs and diseases, which can have serious implications \cite{ghosh2023aclm, hu2023gda}. This limitation seriously affects the performance of LLMs in biomedical NLP tasks, ranging from extraction tasks to question answering task, leading to potential misinterpretations of biomedical documents. Therefore, how to address this data gap is crucial for enhancing the accuracy and reliability of BioNLP applications \cite{xu2022can}.

\begin{figure}
    \centering
    \includegraphics[width=\linewidth]{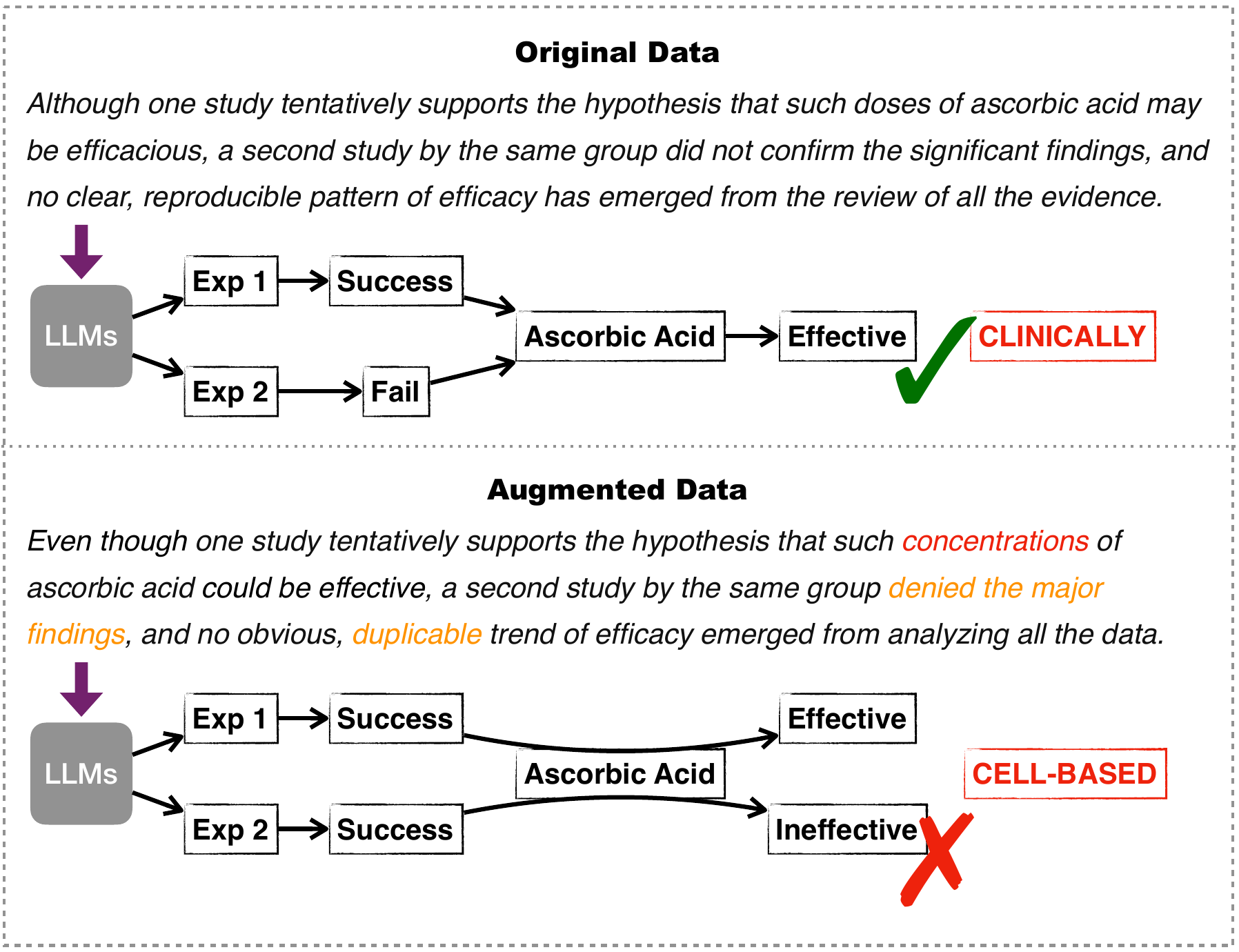}
    \caption{Examples for falsely augmented data. Slightly rephrasing the original sentences with logical errors or mis-replacing similar words leads to a totally wrong models understanding. Further, substituting words from ``dose" to ``concentration" changes the whole sentence's in-context meaning, and damages model wrong understanding from one sentence to the whole passage.}
    \label{fig:demo_for_existing_DA_methods}
\end{figure}

A widely-used technique is to adopt Synthetic Data Augmentation (DA) to solve data scarcity problem in the BioNLP tasks \cite{ghosh2023aclm, hu2023gda}, which can be further divided into two categories: Amendment Rule-based methods and Generation-based methods. The former one utilizes similarity or pre-defined rules amending the given input data to obtain similar instances \cite{zeng2016incorporating, wei2019eda} while the latter one adopts generative models to generate pseudo data based on fixed prompts or instruction \cite{cai2020data, bayer2023data}. Whereas, illustrated by Figure \ref{fig:demo_for_existing_DA_methods}, simply amending "\textit{concentration}" to "\textit{dose}" or paraphrasing "\textit{...did not confirm...}" to "\textit{...denied...}" would cause logical errors which cut off the inner connection between diseases and drugs\footnote{Generally, ``concentration" and ``dose" implies different in-context circumstance. And forward affect description cannot be replaced by negatively reverse description in the biomedical domain.} Those counterfactual data result in the models' misunderstanding among diseases and rugs, and further damage the confidence of biomedical models. Further, both existing DA methods will poison biomedical models.

\begin{figure*}[!t]
    \centering
    \includegraphics[width=\textwidth]{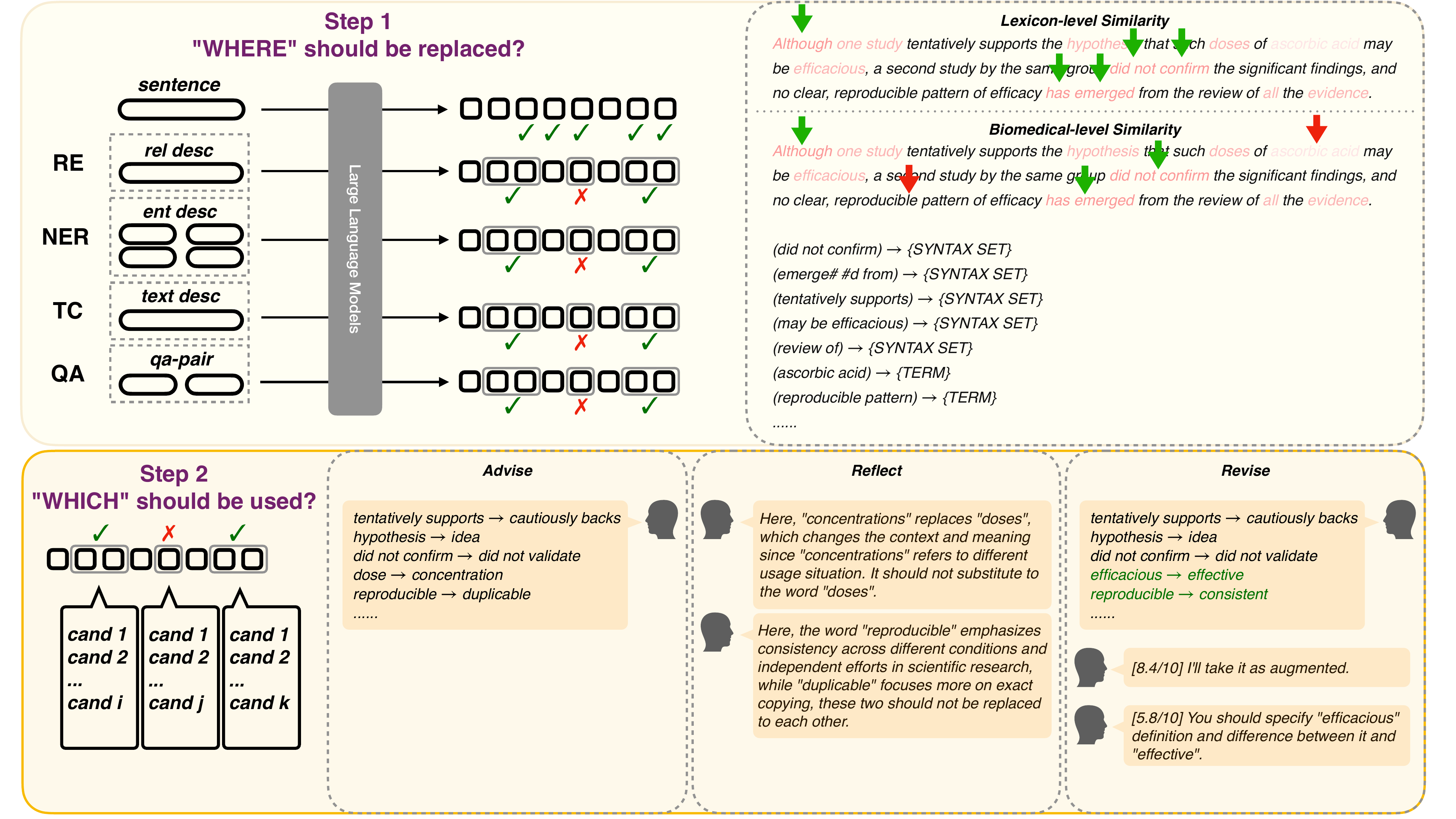}
    \caption{Overview framework of BioRDA. Both lexicon and biomedical-level similarity are evaluated to find the most appropriate position to rephrase in step 1. And step 2 adopts a multi-agent involved system to select the best rephrase candidate.}
    \label{fig:overview_framework}
\end{figure*}

To this end, as shown in Figure \ref{fig:overview_framework}, this paper proposes a Biomedical-dedicated Rationale-based approach, named BioRDA, for biomedical data augmentation. To tackle the wrong paraphrase and mis-replacement issues, we fractionize the process with two self-questions: ``\texttt{WHERE}" and ``\texttt{WHICH}". ``\texttt{WHERE}" stands for finding the replacement position, which must have a strong correlation with the given bio-relation AND diverse lexicon relativity. ``\texttt{WHICH}" indicates selecting the most rational words from word candidates with similar meanings. In the position selection phase, we propose multiple factors involved similarity evaluation method to compute the contribution score of each token in the input sentence. This method considers the similarity of \textbf{token and entities} AND \textbf{token-entity tuples and bio-relation}, to construct the contribution score map. The map represents each replaced token can from what extend affect the whole sentence by finding the trade-off, which is increasing lexicon diversity while under the bio-relation restriction. In the word selection phase, we adopt the ``Advise-Reflect-Revise" paradigm by constructing a multi-agents debate system. At each iteration, the system randomly selects one agent as an Adviser to propose the generation advice. The other agents give the reviews based on word definition, word similarity, syntax correctness, and usage examples. Then the adviser tries to reflect and revise the generated sentence by combining the reviews. Finally, the other agents will give the score based on their own consideration. The augmented sentence will be accepted if the total score is higher than the threshold. In this way, each agent has the same probability to be the adviser and amend the sentence. And the other agents can put reviews on and reflect on how to align the different reviews, to keep the augmented sentence far away from mis-replacement.

We conducted extensive experiments on 9 commonly used BioNLP datasets in 4 different biomedical tasks selected by both BLURB and BigBIO benchmarks. Our proposed BioRDA outperforms all the baseline models by an average of 2.98\%. The experimental results demonstrate that BioRDA is beneficial for alleviating counterfactual issues in four biomedical tasks. Besides, we make a strict analysis of each component of our BioRDA to highlight generating biomedical instances to alleviate the data scarcity problem. 

In summary, the contributions of this paper are as follows:

\begin{itemize}[leftmargin=*]
    \item This paper proposes a biomedical-dedicated rationale-based synthetic data augmentation method for BioNLP tasks. A novel similarity computing approach helps model to augment data with diverse lexicon and under bio-relation restriction. Another multi-agents system promotes the model to escape falling into mis-replacement trap.
    \item This paper demonstrates the benefits of alleviating counterfactual issues of BioNLP models. Our proposed BioRDA outperforms all other baseline models by an average of 2.98\% in four biomedical tasks in BLURB and BigBIO benchmarks.
    \item Additional experiments demonstrate that both \texttt{WHERE} and \texttt{WHICH} information are essential for generating high-quality biomedical instances, with \texttt{WHERE} information helping the model grasp syntax and \texttt{WHICH} information enabling it to accurately distinguish specialized biomedical terms.
\end{itemize}

\section{Related Works}

\subsection{LLMs for Biomedical Tasks}

Though Large Language Models (LLMs) help several Natural Language Processing (NLP) tasks attain promising milestones, due to the scarcity of annotated data and yielding experts to involve their knowledge in data annotation is costly, the key challenge of biomedical tasks is performing better results with the limited well-annotated data \cite{lee2020biobert, tinn2023fine, omiye2024large}. To address this problem, there are three popular techniques adapted to BioNLP tasks, which are domain supervised learning \cite{beltagy2019scibert, lee2020biobert}, indirect supervised learning \cite{roth2017incidental, xu2022can}, and data augmentation \cite{lee2021neural, hu2023gda, ghosh2023aclm}. Domain supervised learning and indirect supervised learning train models on biomedical-related corpus. The former one is aiming to train on the larger data to obtain better performance while the latter is designed to convert the IE task to another formatting task (e.g., Machine Reading Comprehension, Question Answering, and Natural Language Inference, etc.) to use extra supervision signal promoting the performance of IE models \cite{xu2022can}. The aforementioned approaches are facing two challenges: (a) the issue of lacking data still obstructs the models, especially for hardly seen instances; (b) there is a huge gap between biomedical data and pre-trained corpus, hence the model is hard to align external domain knowledge with its intrinsic parameterized knowledge.

\subsection{Synthetic Data Augmentation in BioNLP}

The synthetic data augmentation aims to generate pseudo instances corresponding to the given instances in semantics but with diverse syntax. The existing methods can be divided into two categories: (a) Amendment-based methods. This kind of method tries to amend the exact token or order of input sentences according to pre-defined rules to augment the data. \cite{wei2019eda} adopts synonym replacement, random insertion, random swap, and random deletion to augment the original sentences. Beyond the token-level, \cite{lee2021neural} interpolates the embeddings and labels of two or more sentences from representation-level. and (b) Generation-based methods. These years, leveraging generative large language models to generate kinds of data has become popular. \cite{anaby2020not} and \cite{papanikolaou2020dare} fine-tune multiple generative models for each relation type to generate augmentations. \cite{bayer2023data} proposes a sophisticated generation-based method that generates augmented data by incorporating new linguistic patterns. Beyond that, \cite{openai2023gpt4} introduced ChatGPT to make great progress in nearly all kinds of NLP tasks. It has proved to be effective utilizing ChatGPT as a data augmentation technique to enrich the data instances \cite{van2023improving}.

However, the existing data augmentation methods generally lean to generate instances leaving faithfulness and factuality alone which poison the model in understanding the interaction among biomedical entities and relations, and further, misleading models cannot be adapted to the real scenarios.

\section{Re-build the Logical Coherence}

\subsection{BioNLP tasks Definition}
\label{sec:pro_formu}

The BioNLP tasks aim to analyse biomeical terms in the long passage $p$. The process can be divided into three levels: lexicon-level (NER and RE), sentence-level (TC), and passage-level (QA). The specific task definition is as follows:

\paragraph{Lexicon-level} The \textbf{Bio-Named Entity Recognition} task is designed to extract specific biomedical entities \({\rm ent} \in \mathbb{E}\) by identifying their positions, \({\rm ent}_{\text{start}}\) and \({\rm ent}_{\text{end}}\), and classifying them into predefined entity types. The \textbf{Biomedical Relation Extraction} task aims to predict relations \(r \in \mathbb{R}\) by analyzing a given entity pair \([{\text{ent}}_1, {\text{ent}}_2]\) within a sentence \(s = \{w_i : i = 0, \ldots, n\}\), where \(n\) represents the length of the sentence, and \(w_i\) denotes the \(i\)-th token.

Both tasks require models to comprehend the meanings of biomedical entities and the relationships between them.

\paragraph{Sentence-level} The \textbf{Biomedical Text Classification} task is designed to determine the topic \(t \in \mathbb{T}\) of a given text or passage $p=\{s_i : i=0,\ldots,n\}$. This task requires the model to analyze the content of the entire passage and classify it into one of the predefined topics. And it can be viewed as a higher-level task that evaluates a model’s understanding of the entire texts.

\paragraph{Passage-level} The \textbf{Question Answering} task involves generating a relevant answer \(a \in \mathbb{A}\) to a question \(q \in \mathbb{Q}\) based on a provided passage $p=\{s_i : i=0,\ldots,n\}$. Here, the model must extract and interpret information from the passage to accurately address the question.

These four BioNLP tasks collectively assess the ability of biomedical models to understand three levels information.

\subsection{Synthetic Data Augmentation Formulation}

The Data Augmentation task aims to enhance model performance by generating additional training data through various transformations. For BioNER, augmentation creates or alters biomedical entities \({\text{ent}} \in \mathbb{E}\) while preserving entity positions \({\text{ent}}_{\text{start}}\) and \({\text{ent}}_{\text{end}}\), helping the model generalize to unseen entities. In RE, new sentences or modified entity pairs \([{\text{ent}}_1, {\text{ent}}_2]\) are generated within a sentence \(s = \{w_i : i = 0, \ldots, n\}\), improving the model’s ability to identify relations \(r \in \mathbb{R}\). For TC, augmentation involves paraphrasing passages \(p = \{s_i : i = 0, \ldots, n\}\), enhancing the model's capability to classify topics \(t \in \mathbb{T}\). In QA, new question-answer pairs \((q \in \mathbb{Q}, a \in \mathbb{A})\) or rephrased questions are generated, enabling better extraction of relevant information from passages \(p\).

\subsection{Data Reconstruction}
\label{sec:data_reconstruction}

\begin{figure}[!t]
    \centering
    \includegraphics[width=\linewidth,trim={22cm 11cm 22cm 11cm},clip]{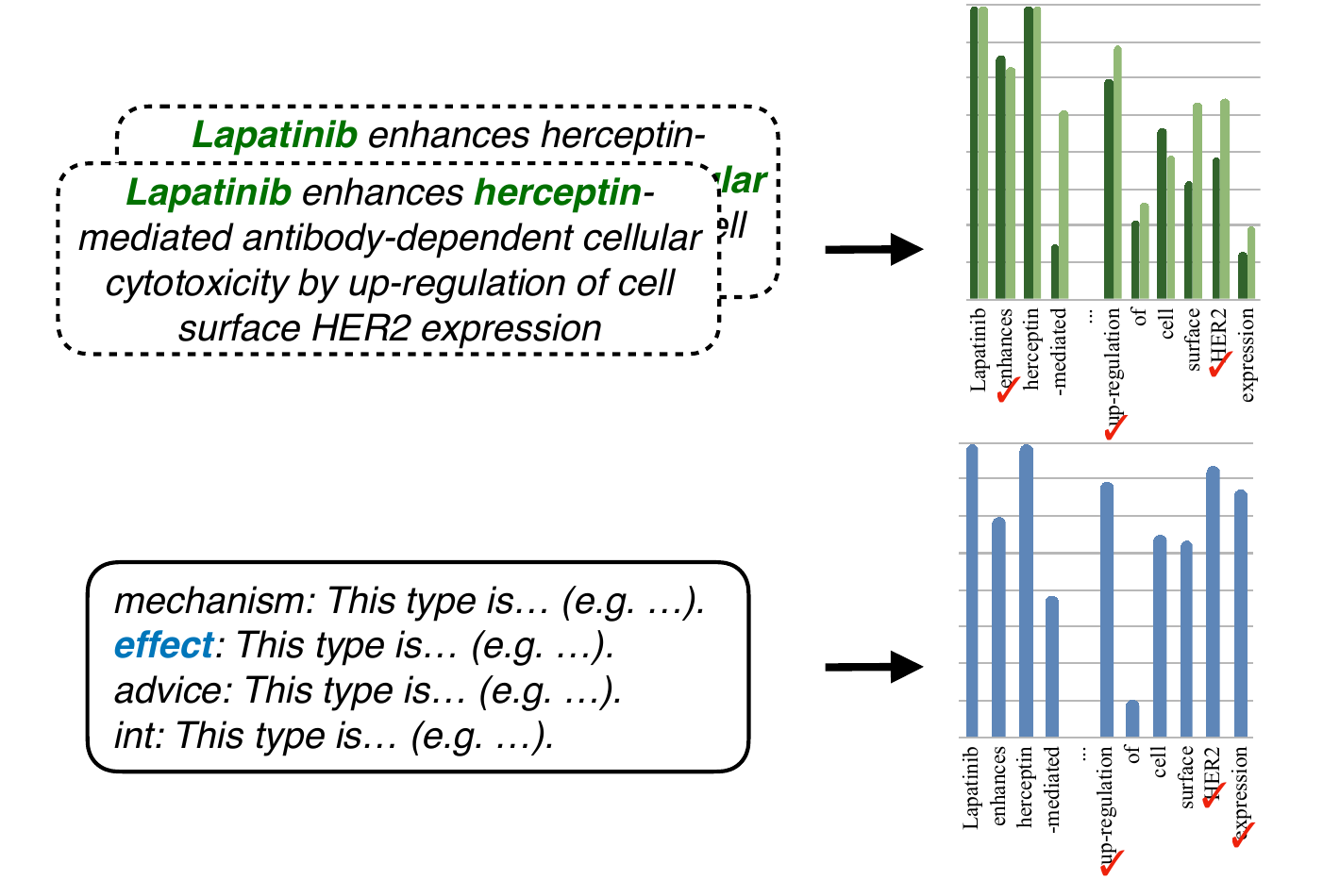}
    \caption{Demo for data reconstruction with lexicon diversity and under relation restriction. The figure on top represents the position that could be replaced by computing lexicon similarity while the bottom one represents the token that is the most relevant to biomedical relations.}
    \label{fig:simi}
\end{figure}

To ensure that augmented instances remain highly relevant to the original biomedical context, the model must effectively know the attribution map of the input text and the broader biomedical domain. This section introduces an Attribution Selector designed to preserve semantic integrity across various BioNLP tasks. The Attribution Selector consists of two key modules: the first leverages lexicon mutual information, denoted as \(\text{attr}_{\text{logits}}\), to introduce diverse lexical information into the augmented data, while the second employs bio-contextual descriptions, denoted as \(\text{attr}_{\text{bio}}\), to ensure that attributions closely align with the specific biomedical context. By doing so, the Attribution Selector enables the model to generate contextually relevant yet diverse instances, thereby maintaining the necessary balance between innovation and domain-specific consistency. For different tasks, the representation has slight divergence. The difference is mainly around the involved elements to compute the similarity for both grammar and relation relativity, which can be referred in Figure \ref{fig:overview_framework}. The parts that need adjustment are the length and specific content involved in the computation, which will not interfere with the overall framework's operation. So in the following methodology introduction, we use BioRE task as an example to illustrate the framework.

\paragraph{Lexicon Diversity} Figure \ref{fig:simi} (a) demonstrate the lexicon diversity similarity computing. For each sentence $s$, we separate the subjective entity and objective entity corresponding to the relation $r$. For each target entity $e=\{w_u,\cdots,w_v\}$, we consider every token except the target entity as candidate keywords $W^*=\{s/e\}$. Then we could know the relativity between every token $w_i\in W^*$ and the target entity $e$ by computing the contribution from token $w_i$ to target entity $e$ denoted as ${\rm attr}(e\leftarrow w_i)$ (i.e., token ``dose" should have closer relativity to the entity "paracetamol" instead of token ``wipes" in clinical documents). Inspired by reformulating input sequence could manipulate the probability distribution of outputs, then affect the prediction result of model \cite{chuang2023dola}. Hence we could give every token the initial score based on what extent to which the token affects the final result, to represent the token itself for the later contribution computing. Here we choose Leave-One-Out (LOO) \cite{lipton2018mythos} as our initial score algorithm denoted as ${\rm attr}(x)$. LOO tries to split each token in the sentence. Then compute the model logits by removing the split token to evaluate each tokens' contribution to the sentence. The lexicon-level attribution algorithm is as follows:
\begin{equation}
\begin{aligned}
    {\rm attr}(e\leftarrow w_i)=&{\rm attr}(e)-{\rm attr}(e\setminus w_i)\\
    {\rm attr}(e\setminus w_i)=&{\rm attr}(s\setminus w_i)-\\
    &{\rm attr}(s\setminus\{w_u,\cdots,w_v,w_i\})
\end{aligned}
\end{equation}
Then we go through all tokens in candidate keywords for all entities and obtain a contribution map corresponding to the target entity. After obtaining the contribution map, as the relation $r^*$ exists between the two related entities ${\rm ent}_1$ and ${\rm ent}_2$, the contribution between these two entities (${\rm attr}(e_1\leftarrow e_2)$ or ${\rm attr}(e_2\leftarrow e_1)$) should be the highest. Hence we fix the contribution map by setting these attributed scores as 1 to be the highest. The others will be scaled in the same proportion. Through this, we ensure the two related entities have the closest relation in the given sentence. The other tokens have their own contribution score indicating to what extent relating to the target entity. And in this way, we have the fixed contribution map with the lexicon-level information, denoted as ${\rm attr}_{\rm logits}=\{(w_i,{\rm attr}(e\leftarrow w_i)):w_i\in \{s\setminus e\}\}$.

\paragraph{Relation Restriction} Figure \ref{fig:simi} (b) shows how relation restriction works. The general paradigm of generating pseudo instances is replacing single or multiple keywords regardless of whether the corresponding bio-relation was. However, these keywords have special meaning to compose the entire sentence which cannot be replaced or need to be regarded as a whole part. For example, the keyword ``\textit{up-regulation...enhance}" means a molecule has positive effectiveness to another molecule and it cannot be placed as ``\textit{downward...not enhance}" which breaks the basic biomedical law. The whole part should be strictly considered with the corresponding relation to avoid making wrong replacements. To overcome this semantic consistency gap between the given sentence and the pseudo sentence, we incorporate bio-relation description into Attribution Selector for selecting semantic-level attribution.

Specifically, according to the datasets selected by BLURB benchmark, each relation type has a specific definition (e.g., relation ``\textit{mechanism}" in DDI dataset is defined as ``\textit{This type is used to annotate DDIs that are described by their PK mechanism}"\footnote{DDI also give an example to clarify each relation. As for ``mechanism", the given example is ``Grepafloxacin may inhibit the metabolism of theobromine" which indicates the entities ``Grepafloxacin" and ``theobromine" have the relation of ``mechanism".}). In this way, we consider this definition as a biomedical supervised signal of the corresponding relation in datasets, denoted as $r_{\rm bio}$ for relation $r$. We then adopt an inference model to evaluate the contribution score between the specific relation added bio-relation description and every token with the entity pair. For each target entity $e^i=\{w_u^i,\cdots,w_v^i\}$, we consider every token except the target entity as candidate keywords $W^*=\{s\setminus \{e^1,e^2\}\}$. For every token $w_i\in W^*$ in candidate keywords, we compute the contribution from $(w_i,e^1,e^2)$ to the relativity between $s$ and $r_{\rm bio}$ as the contribution score ${\rm attr}(r_{\rm bio}\leftarrow w_i)$ as follows:
\begin{equation}
\begin{aligned}
    {\rm attr}(r_{\rm bio}\leftarrow w_i)=&{\rm attr}(e^1,e^2)-{\rm attr}(e^1,e^2\setminus w_i)\\
    {\rm attr}(e^1,e^2\setminus w_i)=&{\rm attr}(s\setminus w_i)-\\
    &{\rm attr}(s\setminus\{e^1,e^2,w_i\})
\end{aligned}
\end{equation}
where ${\rm attr}(x)$ is the same function as Lexicon-level Attribution. To generalize the contribution score, we set the original sentence should be the closest to the relation $r$ with the value of 1. In the same meaning, if the sentence without the entities pairs $e^1$ and $e^2$, the relation has no meaning in this sentence. Hence we set the ${\rm attr}(s\setminus \{e^1,e^2\})$ with the value of 0 indicating that the sentence without the subjective and objective entity should have the lowest relativity with the relation. Then we fix all contribution scores with the same rule as the Lexicon-level Attribution. In this way, we have the fixed contribution map with bio-relation description denoted as ${\rm attr}_{\rm bio}=\{(w_i,attr(r_{\rm bio}\leftarrow w_i)):w_i\in\{s\setminus\{e^1,e^2\}\}\}$

\paragraph{Attribution Mask.} To obtain the keywords that both have various lexicon-level attribution while not leaving far from the specific bio-relation coherence, we combine the two contribution maps ${\rm attr}_{\rm logits}$ and ${\rm attr}_{\rm bio}$ by selecting top-$n$ common keywords with the $n$ highest contribution score. Now we have $n$ keywords set $K$ corresponding to a specific sentence $s$ with subjective and objective entities $E=\{e^1, e^2\}$ representing the relation $r$. Then we mask the other tokens out of the $K$ and the entities $E$. To make the two entities prominent, as shown in Figure \ref{fig:overview_framework}, following the \cite{zhong2020frustratingly}, we concatenate the two entities behind the original sentence and add label marker ahead and after the entity to explicitly indicate the entity type and give the model hint to recognize the start and the end of the entity, denotes as $e_*={\rm\langle s: ent_{type}\rangle} e {\rm\langle /s: ent_{type}\rangle}$. We denote the new input sentence as $X=\{s_{\rm masked} | e^1_* | e^2_*\}$.

\begin{figure*}[!t]
    \centering
    \includegraphics[width=\linewidth,trim={5cm 12cm 5cm 12cm},clip]{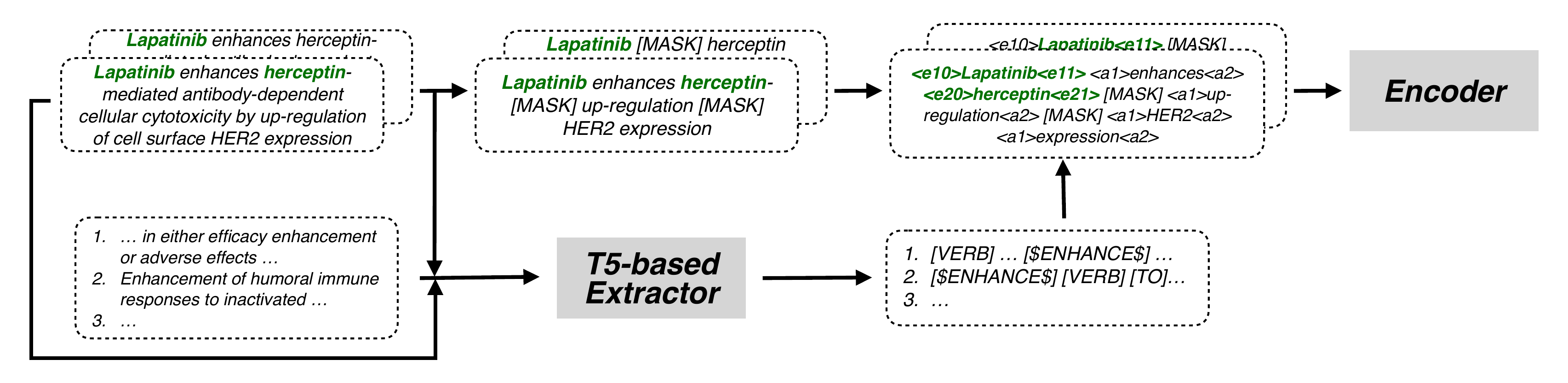}
    \caption{Pseudo instances generation training process. To proceed with the generation, T5 model is applied to learn the WHERE position information while extracting the syntax feature to help model understand the sentence pattern.}
    \label{fig:mask}
\end{figure*}

\subsection{Pseudo Data Generation}
\label{sec:pse_data_gene}

Figure \ref{fig:mask} shows the whole data reconstruction process. The proposed model BioRDA adopts T5 as a pseudo data generator in sequence-to-sequence paradigm. The Encoder is expected to understand the interaction between selective keywords and the subjective and objective entities while also the relation between them. The input $X=\{s_{\rm masked} | e^1_* | e^2_*\}$ is served as template with masked tokens and the BioRDA learns to fill the blank to recover the original sentence. Specifically, the encoder $f_{\rm en}(\cdot)$ takes input tokenized sentence $X=\{x_1,x_2,\dots,x_n\}$ and obtains the contextualized token embeddings $H=f_{\rm en}(X)=\{h_1,h_2,\dots,h_n\}$. The recover decoder $f_{\rm de}(\cdot)$ aims to recover the masked part of the input sentence with the training object $Y$: 
\begin{equation}
    Y=\{y_i: y_i=\max p(y_i|y_{<i},H)\}
\end{equation}

As for augmenting data with the same relation type, a natural paradigm is sampling similar sentences from training set and making the decoder part to learn the pattern from similar sentences to generate different instances with the embedding $H$ from trained encoder. However, we argue that the model frequently imitates a similar sentence by using the common show-up words instead of finding the key structure of the similar sentences. In this way, especially in the biomedical field, the sentence would be performed like piling up plenty of terminology but constructed without biomedical logic. In this paper, we add a syntax re-fix module $f_{\rm ex}(\cdot)$ to help model augment data like transcribing from biomedical materials.

Firstly, we sample another $k$ sentence $\tilde{s}$ with the same subjective and objective entities $e^1$ and $e^2$ with the same relation type $r$. Then we concatenate the above sentence with the target sentence $s$ as the first one, feeding into the T5-based extractor to extract the key structure $\bar{s}$ of the ten sentence. Then we compute the similarity between the $\bar{s}$ and the given sentences concatenated with the biomedical relation notion denoted as $s|r_{\rm bio}$ and $\tilde{s}|r_{\rm bio}$, respectively, till the similarity score is all higher than 80\%. We consider the obtained $\bar{s}$ contains the key structure.

Sequentially, we add bio-relation description not only in attribution selector, but also incorporate it into the decoder, to enhance the decoder the ability to generate more biomedically logical pseudo instances, we feed encoder's output $H$, and target biomedical relation notion $r_{\rm bio}$ and the biomedical key structure $\bar{s}$ to the decoder to generate pseudo instances. In this way, the decoder can used to augment the sentence by maximizing $p(y_i|y_{<i},H,r_{\rm bio},\bar{s})$ to obtain the pseudo sentences $s_{\rm aug}$:
\begin{equation}
    Y=\{y_i: y_i=\max p(y_i|y_{<i},H,r_{\rm bio},\bar{s})\}
\end{equation}

Overall, instead of amending the original sentence purely based on replacing each token with the similar one, the model could be re-built the logical coherence by considering both lexicon and semantic similarity to distinguish which token has the closest relativity to the subject and object entities pair, while contributes the bio-relations most. Besides, because of the syntax re-fix module, the augmented sentence is more like the standard biomedical one rather than based on general domain.

\section{Fine-Grained Reflection}

Through the Attribution Selector mechanism, the model could explicitly amend the original sentence regarding the relativity of entities pair and bio-relation. However, in biomedical domain, similar words may have totally different usage and meaning. For example, ``dose" and ``concentration" can be the similar instructions around what extent the medicine should be given to the acceptor. However the word ``concentration" generally implies the context is a cell-based instruction of outside body in laboratory which may not be directly applied in clinical situation. While the word ``dose" is more commonly used in clinical records. Hence accurate word usage can be vital in biomedical domain, especially in long-context documents, trivial discrepancies would cause misunderstanding of the whole document. In this case, we propose a Fine-Grained Reflection mechanism with multi-agents to make a explicit and reasonable distinction among candidate tokens. Figure \ref{fig:overview_framework} clearly demonstrates the multi-agents system.

Before going through the Fine-Grained Reflection, we have the original sentence $s$ and augmented sentence $s^*$. Regarding the massive pre-trained corpus and tremendous parameters, large language models were proved to have strong ability to recognize and discriminate different input sequences. Here we set $n$ LLM-based agents, and at each iteration, randomly select one as judge agent ${\rm Agent_{j}}$ to discriminate the original $s$ and augmented $s^*$ by asking it to perform the reviews about greatest discrepancy between these two sentences. Each discrepancy review between $s$ and $s^*$ is a tuple showing what is like in $s$ and $s^*$ respectively. Then the other $n-1$ agents need to elaborate whether this amendment is in reason from word definition, word similarity, syntax correctness, and using example\footnote{Prompts and detailed reasoning description are in Section \ref{apd:prompts}}. Then ${\rm Agent_{j}}$ should take amendment reasons and refine the augmented sentences $s^*$. At each iteration, all other agents should grade the new augmented sentence $s^*$ to be the acceptance score of augmented sentence. If the acceptance score is lower than the threshold, the original $s$ and this new augmented $s^*$ should feed to the LLM-based agents in next iteration. The detailed algorithm is demonstrated in Algorithm \ref{alg:alg1}. In this way, each agent has the same probability of being the judge agent and amending the sentence. And the other agents can put reviews on and reflect on how to align the different reviews, to keep the augmented sentence far away from mis-replacement.

\begin{algorithm}
    \caption{Fine-Grained Reflection}
    \begin{algorithmic}
        \REQUIRE $n$ LLM-based agents, threshold $\sigma$ for acceptance score
        \STATE Given original sentence $s$ and augmented sentence $s^*$
        \WHILE{acceptance score of $s^*\leq\sigma$}
        \STATE Randomly select one agent as judge agent ${\rm Agent_{j}}$
        \STATE ${\rm Agent_{j}}$ reviews and identifies discrepancies between $s$ and $s^*$
        \STATE Discrepancy review is a tuple showing differences in $s$ and $s^*$
        \FOR{each of the other $n-1$ agents}
            \STATE Elaborate on whether the amendment is reasonable
        \ENDFOR
        \STATE ${\rm Agent_{j}}$ refines the augmented sentence $s^*$ based on the amendment reasons
        \FOR{each of the other agents}
            \STATE Grade the new augmented sentence $s^*$ to determine acceptance score
        \ENDFOR
        \IF{acceptance score is below threshold}
        \STATE Feed the original $s$ and the new augmented $s^*$ to the LLM-based agents in the next iteration
    \ENDIF
\ENDWHILE
    \end{algorithmic}
    \label{alg:alg1}
\end{algorithm}

\section{Experiments}

We conduct extensive experiments across three key biomedical tasks: relation extraction (bioRE), named entity recognition (bioNER), and text classification (bioTC). To evaluate our proposed BioRDA, we select 9 datasets from both the BLURB and BigBIO benchmarks, including BC5, NCBI, BC2GM, JNLPBA, ChemProt, DDI, GAD, PubMedQA, and HoC. Our analysis demonstrates how BioRDA enhances the model's ability to understand complex relationships, such as disease-drug interactions and drug-drug interactions, even in low-resource settings.

\subsection{Experimental Settings}

\paragraph{Datasets} BLURB and BigBIO are two commonly used BioNLP benchmark containing multiple biomedical relevant tasks targeting enhance models' ability to understand relation among biomolecules. Here we choose 4 key biomedical tasks, BioRE, BioNER, BioQA, and BioTC, to verify our proposed data augmentation methods. (1) \underline{NER}: \textbf{BC5} is designed for the extraction of chemical-disease relations. It consists of 1,500 PubMed articles with annotations for 4,409 chemicals, 5,818 diseases, and 3,116 chemical-disease interactions \cite{li2016biocreative}. \textbf{NCBI} Disease Corpus contains 793 PubMed abstracts and focuses on disease mention recognition, providing a detailed annotation of disease names and their variants \cite{dougan2014ncbi}. \textbf{BC2GM} was created for gene mention recognition, comprising 20,000 sentences from PubMed abstracts, annotated with gene and protein names \cite{smith2008overview}. \textbf{JNLPBA} was used for the shared task of identifying named entities in biomedical texts, particularly focused on entities like proteins, DNA, RNA, cell types, and cell lines \cite{collier2004introduction}. (2) \underline{RE}: \textbf{GAD} is a semi-labeled dataset created using Genetic Association Archive and consists of gene-disease associations. The purpose of the dataset is to determine whether there was a gene-disease relationship between the given subjective and objective entities which is a bi-label prediction \cite{becker2004genetic}. \textbf{DDI} is also named Drug-Drug Interaction indicating the five pre-defined interaction types between different drugs which was specialized in pharmacovigilance built from PubMed abstracts \cite{herrero2013ddi}. \textbf{ChemProt} is a disease chemical biology database, which is based on a compilation of multiple chemical–protein annotation resources, as well as disease-associated protein-protein interactions (PPIs) \cite{taboureau2010chemprot}. \underline{QA}: \textbf{PubMedQA} dataset is a biomedical question-answering dataset that contains approximately 1,000 annotated question-answer pairs and an additional large-scale, unannotated dataset of around 211,000 QA pairs \cite{jin2020pubmedqa}. \underline{TC}: \textbf{HoC} consists of 1852 PubMed publication abstracts manually annotated by experts according to the Hallmarks of Cancer taxonomy. The taxonomy consists of 37 classes in a hierarchy. Zero or more class labels are assigned to each sentence in the corpus \cite{baker2015automatic}.

\paragraph{Baselines} We use different size-scale models for two phases of our proposed method. In grammar and bio-relation similarity computing phase, we fine-tune the relatively small pre-trained language models to build the reasoning chain forest. And in the agent discussion phase, we apply multiple large language models to construct self-debate system to generate reasoning, reflecting, and distinguishing thought chains. And finally generate the accurate augmented data instances. As for small pre-trained language models, we categorized them followed by \cite{xu2022can}: (1) Pre-trained on Semantic Scholar: \textbf{SciBERT} \cite{beltagy2019scibert} is pre-trained on BERT-based model by providing specific prompt for each relation. The training corpus consists of academic papers selected from Semantic Scholar. (2) Pre-trained on PubMed articles: \textbf{BioBERT} \cite{lee2020biobert} was pre-trained on a commercial-collection subset of PMC while \textbf{BioLinkBERT} \cite{yasunaga2022linkbert} adopt link prediction task in pre-training processing which helped model to learn the multi-hop inference. And \textbf{PubMedBERT} captures the unique language, terminology, and nuances of biomedical texts more effectively than general-purpose language models on PubMed articles. (3) Beyond, we also compare our BioRDA with an indirect supervision method \textbf{NBR-NLI} \cite{xu2022can} which converts the relation extraction task to NLI formulation trained on the BioLinkBERT-large backbone. We adopt the version of NBR$_{\rm NLI+FT}$ which was fine-tuned on two general domain NLI datasets retaining biomedical domain knowledge and learning relevant NLI knowledge.  As for large language models, we apply \textbf{LLaMA2} and \textbf{ChatGLM} to build the agent discussion system. The two LLMs are designed for general-purpose NLP tasks which have strong ability on linguistic pattern mining and reasoning. To prove effectiveness of our proposed method, we select the highest 3 backbone models for all NER, RE, QA, and TC tasks commonly on BLURB and BigBIO benchmark\footnote{Please refer to \href{https://paperswithcode.com/dataset/blurb}{BLURB} and \href{https://paperswithcode.com/paper/bigbio-a-framework-for-data-centric}{BigBIO}.}. And for each specific sub-tasks, we find the most comparable SoTA model to verify its effectiveness.

\paragraph{Experimental Implementation} For all the models, we use comparable hyper-parameters and evaluation metrics. Specifically, for small pre-trained language models, we fine-tuned them mainly on  NVIDIA(INNO3D) GeForce RTX 2080Ti. For large language models, we mainly conduct our experiments on two different backbone models: llama2-chat-7b, chatglm3-10b. And GPT-4 was applied as judge to evaluate the generation quality and give the score with multiple pre-defined criteria, following lots of previous works \cite{achiam2024gpt4, koutcheme2024opensource, openai2023gpt4}. For the evaluation metrics, we strictly follow the BLURB and BigBIO benchmarks, we use F1 entity-level for BioNER, Micro F1 for BioRE, and Average Micro F1 for BioTC. For all large language models, we set all the temperatures and top p as 0.1 to reduce the randomness of LLMs for all methods, rendering a more fair comparison. We access proprietary models through Azure API and deploy open-source models through FastAPI. The detailed prompts for all model usage and versions of models can be found in Section \ref{apd:prompts}.

\begin{table*}[!t]
    \centering
    \caption{Detailed comparison of experimental results across nine datasets spanning four tasks. The \colorbox{lightgray}{Gray} cells show the best DA method performance. Ave column indicates average performance across all tasks regardless of the original metric.}
    \aboverulesep=0ex
    \belowrulesep=0ex
    \begin{tabular}{c|cccc|ccc|c|c|cc}
        \toprule
        \rule{0pt}{10pt}
        \multirow{2}{*}{Model} & \multicolumn{4}{c|}{NER} & \multicolumn{3}{c|}{RE} & QA & TC & \multirow{2}{*}{Ave} & \multirow{2}{*}{$\Delta$\%}\\
        \rule{0pt}{8pt}
        & BC5 & NCBI & BC2GM & JNLPBA & GAD & DDI & ChemProt & PubMedQA & HoC &&\\
        \midrule
        \rule{0pt}{10pt}
        \textbf{PubMedBERT}            & 89.48 & 87.82 & 84.52 & 79.10 & 82.34 & 82.36 & 77.24 & 55.84 & 82.32 & 80.11\\
        w/ EDA                  & 89.62 & 87.93 & 84.71 & 79.28 & 82.55 & 82.49 & 77.47 & 56.03 & 82.47 & 80.28 & 0.21\\
        w/ ParaGraph            & 89.60 & 88.00 & 84.90 & 79.40 & 82.70 & 82.70 & 77.60 & 56.20 & 82.70 & 80.42 & 0.39\\
        w/ AdMix                & 89.70 & 88.10 & 85.10 & 79.60 & 82.90 & 82.90 & 77.80 & 56.40 & 82.90 & 80.60 & 0.61\\
        w/ LAMBADA              & 89.75 & 88.12 & 85.01 & 79.57 & 82.77 & 82.75 & 77.72 & 56.29 & 82.77 & 80.53 & 0.52\\
        w/ ChatGPT-4o-mini      & 89.94 & 88.37 & 85.48 & 79.98 & 83.17 & 83.14 & 78.11 & 56.73 & 83.21 & 80.90 & 0.99\\
        w/ BioRDA-biolink-llama & \cellcolor{lightgray}90.16 & \cellcolor{lightgray}90.17 & \cellcolor{lightgray}87.10 & \cellcolor{lightgray}81.47 & \cellcolor{lightgray}85.08 & \cellcolor{lightgray}85.10 & \cellcolor{lightgray}79.52 & \cellcolor{lightgray}57.38 & \cellcolor{lightgray}84.85 & \cellcolor{lightgray}82.31 & \cellcolor{lightgray}2.75\\
        \midrule
        \rule{0pt}{10pt}
        \textbf{BioLinkBERT-base}      & 89.93 & 88.18 & 84.90 & 79.03 & 84.39 & 82.72 & 77.57 & 70.20 & 84.35 & 82.36\\
        w/ EDA                  & 90.01 & 88.26 & 85.02 & 79.15 & 84.55 & 82.86 & 77.69 & 70.34 & 84.49 & 82.49 & 0.16\\
        w/ ParaGraph            & 90.12 & 88.36 & 85.17 & 79.28 & 84.67 & 82.97 & 77.84 & 70.49 & 84.61 & 82.61 & 0.30\\
        w/ AdMix                & 90.20 & 88.44 & 85.29 & 79.41 & 84.78 & 83.08 & 77.95 & 70.63 & 84.74 & 82.72 & 0.44\\
        w/ LAMBADA              & 90.32 & 88.53 & 85.44 & 79.55 & 84.91 & 83.21 & 78.08 & 70.76 & 84.89 & 82.84 & 0.58\\
        w/ ChatGPT-4o-mini      & 90.52 & 88.72 & 85.69 & 79.79 & 85.16 & 83.46 & 78.35 & 71.04 & 85.16 & 83.10 & 0.90\\
        w/ BioRDA-biolink-llama & \cellcolor{lightgray}92.60 & \cellcolor{lightgray}90.82 & \cellcolor{lightgray}87.45 & \cellcolor{lightgray}81.40 & \cellcolor{lightgray}86.92 & \cellcolor{lightgray}85.14 & \cellcolor{lightgray}79.89 & \cellcolor{lightgray}72.35 & \cellcolor{lightgray}86.88 & \cellcolor{lightgray}84.69 & \cellcolor{lightgray}2.83\\
        \midrule
        \rule{0pt}{10pt}
        \textbf{BioLinkBERT-large}     & 90.22 & 88.76 & 85.18 & 80.06 & 84.90 & 83.35 & 79.98 & 72.20 & 88.10 & 83.66\\
        w/ EDA                  & 90.34 & 88.82 & 85.35 & 80.21 & 85.07 & 83.48 & 80.13 & 72.35 & 88.26 & 83.78 & 0.14\\
        w/ ParaGraph            & 90.45 & 88.91 & 85.47 & 80.34 & 85.19 & 83.61 & 80.28 & 72.49 & 88.38 & 83.96 & 0.36\\
        w/ AdMix                & 90.54 & 88.99 & 85.61 & 80.46 & 85.32 & 83.74 & 80.42 & 72.62 & 88.52 & 84.06 & 0.49\\
        w/ LAMBADA              & 90.66 & 89.09 & 85.75 & 80.60 & 85.46 & 83.87 & 80.55 & 72.77 & 88.65 & 84.26 & 0.72\\
        w/ ChatGPT-4o-mini      & 91.89 & 90.28 & 87.01 & 81.86 & 86.72 & 85.14 & 81.82 & 73.06 & 89.93 & 85.30 & 1.96\\
        w/ BioRDA-biolink-llama & \cellcolor{lightgray}92.90 & \cellcolor{lightgray}91.43 & \cellcolor{lightgray}87.85 & \cellcolor{lightgray}82.40 & \cellcolor{lightgray}87.45 & \cellcolor{lightgray}85.89 & \cellcolor{lightgray}82.49 & \cellcolor{lightgray}74.30 & \cellcolor{lightgray}90.60 & \cellcolor{lightgray}86.15 & \cellcolor{lightgray}2.98\\
        \bottomrule
    \end{tabular}
    \label{tab:main}
\end{table*}

\begin{table*}[!t]
    \centering
    \caption{Experiments on task-oriented models. UniversalNER, NBR-NLI, and BioGPT are designed for specific biomedical tasks. Vanilla indicates the re-implement experimental results. \colorbox{lightgray}{Gray} cells show the best DA method performance.}
    \aboverulesep=0ex
    \belowrulesep=0ex
    \begin{tabular}{c|cccc|ccc|cc}
        \toprule
        \rule{0pt}{10pt}
        \multirow{2}{*}{Model} & \multicolumn{4}{c|}{\textbf{UniversalNER (NER)}} & \multicolumn{3}{c|}{\textbf{NBR-NLI (RE)}} & \multicolumn{2}{c}{\textbf{BioGPT-large (QA+TC)}}\\
        \rule{0pt}{8pt}
        & BC5 & NCBI & BC2GM & JNLPBA & GAD & DDI & ChemProt & PubMedQA & HoC\\
        \midrule
        \rule{0pt}{10pt}
        Vanilla            & 89.34 & 86.96 & 82.42 & 77.54 & 85.86 & 84.66 & 80.54 & 81.00 & 85.12\\
        w/ EDA             & 89.45 & 87.10 & 82.59 & 77.69 & 85.97 & 84.80 & 80.71 & 81.13 & 85.19\\
        w/ ParaGraph       & 89.54 & 87.22 & 82.72 & 77.81 & 86.06 & 84.91 & 80.83 & 81.19 & 85.19\\
        w/ AdMix           & 89.63 & 87.31 & 82.85 & 77.92 & 86.14 & 85.01 & 80.94 & 81.25 & 85.24\\
        w/ LAMBADA         & 89.72 & 87.41 & 82.97 & 78.04 & 86.22 & 85.11 & 81.05 & 81.20 & 85.59\\
        w/ ChatGPT-4o-mini & \cellcolor{lightgray}89.90 & \cellcolor{lightgray}87.61 & \cellcolor{lightgray}83.22 & \cellcolor{lightgray}78.29 & \cellcolor{lightgray}86.38 & \cellcolor{lightgray}85.31 & \cellcolor{lightgray}81.29 & \cellcolor{lightgray}81.79 & \cellcolor{lightgray}86.46\\
        \bottomrule
    \end{tabular}
    \label{tab:task-ori}
\end{table*}

\subsection{Experimental Results}

Table \ref{tab:main} demonstrates the comparison of experimental results. Our results revealed a substantial improvement in model performance when utilizing our proposed method, particularly in terms of F1 score, accuracy, and precision-recall metrics. The baseline average F1 score for BioLinkBERT-large across all datasets was 83.66\%. After applying our proposed data augmentation techniques, the overall score increased to 86.15\%, a significant 2.98\% point improvement. This enhancement was particularly pronounced in the NCBI dataset, where the F1 score rose from 88.76\% to 91.43\%, marking a 2.67 percentage point gain. Similarly, PubMedBERT and BioLinkBERT-base showed improvements of 2.75 and 2.83 percentage points, respectively, when augmented with our techniques, achieving final overall scores of 82.31\% and 84.69\%.

The performance gains for NER and RE tasks reveal a consistent upward trend, with significant improvements observed across the board. Our model, in particular, demonstrates the highest average performance gains, most notably in NER and RE tasks, where it outperforms other models by a substantial margin. For instance, when applied to BioLinkBERT-large, our BioRDA model achieves an average performance improvement of 2.98\%, which is the highest observed across all configurations. This suggests that the integration of BioRDA into existing models can substantially enhance their ability to process and understand complex biomedical texts. Additionally, the results indicate that while performance improvements are generally robust across tasks, the extent of these gains varies depending on the model and augmentation strategy used. Our model, when applied to BioLinkBERT-large, also shows exceptional performance in QA tasks, particularly in the PubMedQA dataset, where the complexity of the task is notably higher. This further underscores the effectiveness of our approach in tackling more challenging NLP tasks within the biomedical domain. The experimental results affirm that our BioRDA model offers significant enhancements over traditional and even other enhanced models, leading to the highest average performance improvements across the tasks evaluated.

Beyond that, the integration of BioRDA into task-oriented models results in consistent performance improvements across all evaluated tasks. Table \ref{tab:task-ori} shows the experimental results on task-oriented models. The most substantial gains are observed when the models are applied to tasks they were not originally designed for, highlighting the method’s ability to generalize and enhance task-specific models for broader applications. This is particularly important in the biomedical domain, where models need to handle a wide range of tasks, from entity recognition to complex question answering, with high accuracy and reliability. Even for models like UniversalNER, NER-NLI, and BioGPT which are specialized for specific tasks, the application of our BioRDA method significantly expands their capabilities, allowing them to perform well across a diverse set of biomedical NLP challenges.

\subsection{The Right WHERE Information Matters}

\begin{table}[!t]
    \centering
    \caption{Experiments on small PLMs evaluation. We use PLMs here to generate the \texttt{WHERE} information. \textbf{Bold} indicates the best performance of corresponding PLM.}
    \begin{tabular}{cccccc}
        \toprule
        Model & NER & RE & QA & TC & Average\\
        \midrule
        BioBERT-base      & 89.47 & 85.29 & 73.88 & 83.33 & 82.99\\
        SciBERT-base      & 89.88 & 87.59 & 80.49 & 81.12 & 84.77\\
        PubMedBERT        & 97.29 & 92.31 & \textbf{90.52} & 91.22 & 92.84\\
        BioLinkBERT-base  & 94.68 & 88.85 & 85.16 & 88.53 & 89.56\\
        BioLinkBERT-large & \textbf{99.71} & \textbf{98.68} & 86.73 & \textbf{97.79} & \textbf{95.73}\\
        \bottomrule
    \end{tabular}
    \label{tab:plms}
\end{table}

\paragraph{Which Small PLMs Can Find WHERE Best?} Table \ref{tab:plms} revealed that the models under consideration exhibited consistent performance trends across most tasks, with BioLinkBERT-large demonstrating superior syntactic proficiency. This model consistently outperformed its counterparts, achieving the highest scores in accuracy and F1 metrics, which indicates its exceptional ability to discern and utilize syntactic structures effectively. For instance, in the NER and RE tasks, BioLinkBERT-large recorded average F1 scores of 99.71\% and 98.68\%, respectively, significantly outperforming PubMedBERT, SciBERT, and BioBERT. These results underscore the model's robustness in identifying key entities, extracting relationships, and accurately interpreting syntactic nuances in biomedical texts.

However, an exception to this pattern was observed in the Question Answering (QA) task, where PubMedBERT outperformed BioLinkBERT-large, achieving the highest accuracy of 90.52\%, compared to BioLinkBERT-large’s 86.73\%. This deviation suggests that while BioLinkBERT-large excels in tasks requiring deep syntactic understanding, PubMedBERT has a particular strength in navigating the complexities of question-answering scenarios. PubMedBERT was designed and trained on PubMedQA dataset directly with a strong focus on syntactic and semantic comprehension, likely providing it with an edge in QA tasks, where understanding the nuances of questions and providing precise answers is critical. In contrast, BioLinkBERT-large, while still highly competitive in the QA task, maybe slightly less optimized for the specific demands of this task, which often involves intricate question structures and requires a different approach to syntactic analysis.

These findings demonstrate that while training on a comprehensive dataset across various domains can enhance the model's ability to understand and identify where lexical replacements can be made or to discern subtle syntactic differences between different datasets, multitask training can also introduce interpretative biases. This may result in the model underperforming on domain-specific data compared to models that have been exclusively trained for that particular domain.

\begin{table}[!t]
    \centering
    \caption{Experiments on LLMs under different number of \texttt{WHERE} information. Prop indicates the proportion for data with WHERE information of all data. \textbf{Bold} denotes the best performance.}
    \begin{tabular}{cccccc}
        \toprule
        Model & Prop & NER & RE & QA & TC\\
        \midrule
        \multirow{6}{*}{LLaMA} & 0 & 68.23 & 69.57 & 67.89 & 70.12\\
        & 20  & 71.56$\pm$.38 & 72.44$\pm$.97 & 70.11$\pm$.56 & 73.21$\pm$.75\\
        & 40  & 75.89$\pm$.61 & 76.34$\pm$.48 & 72.05$\pm$.22 & 77.77$\pm$.13\\
        & 60  & 77.91$\pm$.03 & 79.12$\pm$.45 & 71.89$\pm$.67 & 81.70$\pm$.64\\
        & 80  & 80.18$\pm$.42 & 81.43$\pm$.11 & 77.12$\pm$.54 & 83.27$\pm$.95\\
        & all & 88.12$\pm$.47 & 89.34$\pm$.89 & 78.90$\pm$.24 & 90.56$\pm$.63\\
        \midrule
        \multirow{6}{*}{ChatGLM} & 0 & 66.27 & 68.45 & 65.31 & 69.19\\
        & 20  & 70.53$\pm$.24 & 71.68$\pm$.35 & 68.77$\pm$.16 & 72.19$\pm$.88\\
        & 40  & 71.12$\pm$.37 & 72.75$\pm$.88 & 68.23$\pm$.75 & 74.90$\pm$.55\\
        & 60  & 75.62$\pm$.53 & 76.81$\pm$.29 & 72.45$\pm$.17 & 77.96$\pm$.61\\
        & 80  & 82.46$\pm$.79 & 84.03$\pm$.57 & 74.25$\pm$.50 & 86.92$\pm$.11\\
        & all & 90.59$\pm$.85 & 91.73$\pm$.14 & 87.49$\pm$.26 & 93.85$\pm$.92\\
        \bottomrule
    \end{tabular}
    \label{tab:where}
\end{table}

\paragraph{Whether WHERE Helps?} Besides, table \ref{tab:where} demonstrates results by adding different proportions of WHERE information, ranging from none to substituting all instances with WHERE information. A dataset comprising 1,000 instances was employed for each task. We systematically introduced WHERE information into the dataset, varying the proportion from 0\% (none) to 20\%, 40\%, 60\%, 80\%, and 100\% (all).

Our findings consistently demonstrate that the inclusion of WHERE information significantly enhances the model's accuracy across all four tasks. For example, in the NER task, the accuracy improved from 68,23\% at 0\% WHERE information to 88.12\% at 100\%. Similarly, in the RE task, the model's performance increased from 69.57\% to 89.34\% as the proportion of WHERE information rose. The QA task exhibited an improvement in exact match scores from 67.89\% to 78.90\%, while in the TC task, accuracy saw an increase from 70.12\% to 90.56\%.

These results underscore the critical role of WHERE information in enhancing a model's contextual understanding, thereby enabling more precise token replacement decisions. The consistent improvement across varying proportions of WHERE information highlights its importance in complex biomedical NLP tasks, where understanding the specific location within a sentence can lead to more accurate and reliable model outputs. This suggests that the integration of location-based information should be considered a key strategy in the development of more sophisticated and context-aware biomedical models.

\subsection{Agent Debate System Selection}

\begin{table}[!t]
    \centering
    \caption{Experiments on Different LLMs to evaluate the instruction following ability. \textbf{Bold} denotes the better LLM performance for corresponding task.}
    \begin{tabular}{ccccccc}
        \toprule
        && word def & word sim & syn & emp & ave\\
        \midrule
        \multirow{4}{*}{LLaMA} & NER & 85.62 & 94.26 & 90.98 & 88.98 & \textbf{89.96}\\
                               & RE  & 89.02 & 90.62 & 80.31 & 94.55 & \textbf{88.62}\\
                               & QA  & 88.89 & 80.70 & 89.11 & 82.56 & \textbf{85.31}\\
                               & TC  & 82.34 & 82.34 & 80.87 & 92.99 & 84.64\\
        \midrule
        \multirow{4}{*}{ChatGLM} & NER & 86.84 & 91.78 & 83.00 & 87.71 & 87.33\\
                                 & RE  & 84.56 & 87.87 & 86.48 & 84.37 & 85.82\\
                                 & QA  & 92.49 & 83.19 & 82.73 & 82.75 & 85.29\\
                                 & TC  & 89.18 & 82.09 & 84.38 & 85.50 & \textbf{85.29}\\
        \bottomrule
    \end{tabular}
    \label{tab:llm}
\end{table}

Table \ref{tab:llm} shows the results of the models' ability to distinguish the differences between augmented instances and original ones. We focused on four key aspects: word definition, word-word similarity, syntax correctness, and word examples\footnote{The prompts and description of verifying these four aspects are shown in Section \ref{apd:prompts}}.

Across nearly all tasks, llama2 consistently outperformed chatglm, demonstrating better performance in distinguishing word meanings, identifying word-word similarities, and maintaining syntactic accuracy. While both models showed competence in handling biomedical language, llama2's consistent superiority across multiple aspects underscores its reliability and effectiveness in biomedical NLP tasks. This comprehensive analysis highlights llama2 as the more robust model, capable of handling the nuanced demands of biomedical language processing with greater precision and accuracy.

\subsection{Generative Instances Quality Analysis}

\begin{table}[!t]
    \centering
    \caption{Experiments on generation quality evaluation. GPT-4 is adopted as judge to give the mark for each biomedical tasks. \textbf{Bold} indicates the highest score.}
    \begin{tabular}{cccccc}
        \toprule
        \textbf{BioLinkBERT-large} & NER & RE & QA & TC & Ave\\
        \midrule
        w/ EDA                  & 85.51 & 85.10 & 85.14 & 85.39 & 85.29\\
        w/ ParaGraph            & 86.50 & 87.12 & 86.19 & 86.30 & 86.53\\
        w/ AdMix                & 87.50 & 87.92 & 86.91 & 87.33 & 87.42\\
        w/ LAMBADA              & 89.53 & 88.14 & 88.07 & 89.39 & 88.78\\
        w/ ChatGPT-4o-mini      & 91.52 & 90.14 & 90.04 & 91.39 & 90.77\\
        w/ BioRDA-biolink-llama & \textbf{94.87} & \textbf{93.49} & \textbf{93.25} & \textbf{94.52} & \textbf{94.03}\\
        \bottomrule
    \end{tabular}
    \label{tab:my_label}
\end{table}

To assess the quality of augmented instances generated by LLMs, GPT-4 was utilized for its fair and professional evaluation of linguistic correctness and overall coherence. This analysis focused on different data augmentation methods applied to the BioLinkBERT-large model. Overall, BioRDA emerged as the superior augmentation technique across the majority of tasks, demonstrating a clear advantage in generating high-quality, linguistically accurate sentences. This method's ability to maintain contextual and semantic integrity highlights its effectiveness for improving the quality of augmented data in biomedical applications. The evaluation revealed that the BioRDA method consistently produced higher-quality sentences compared to other augmentation techniques. In the NER task, sentences augmented using BioRDA were marked as linguistically correct 94.87\% of the time, whereas those generated by the EDA method achieved a correctness rate of 85.51\%. This result underscores the effectiveness of BioRDA in maintaining the linguistic integrity and contextual relevance necessary for accurate entity recognition in biomedical texts. Similarly, in the RE task, the Paragraph-based augmentation method yielded a correctness rate of 87.12\%, while EDA achieved only 85.1\%. And our proposed BioRDA could also achieve 93.49\% beating all other baselines.

\section{Conclusions}

This paper addresses a critical challenge in Biomedical Natural Language Processing (BioNLP) by introducing a rationale-based data augmentation approach specifically designed for biomedical tasks. By integrating a rationale-based framework with a multi-agent system for augmentation, the approach not only enhances data quality but also ensures that augmented instances remain relevant and coherent within the biomedical domain. Our extensive experiments demonstrate the efficacy of this method across various BioNLP tasks. Over four tasks, the proposed BioRDA method outperforms existing baseline models by 2.98\%, validating its capability to alleviate counterfactual issues and improve model performance. The application of our approach to multiple BioNLP tasks further confirms its versatility and effectiveness, as evidenced by the improved results across nine datasets from the BLURB and BigBIO benchmarks.

\section{Prompts of Large Language Models}\label{apd:prompts}

\begin{center}\textbf{--- Prompt 1 ---}\end{center}

\textbf{``Task: Multi-agents System Debate}

\textbf{Initial Statement:}\\
\textit{You are the Lead Agent tasked with presenting an argument on the topic: ‘\textbf{[Insert topic here]}’. Please construct a clear and well-supported statement presenting your point of view. Include relevant examples, evidence, and reasoning to back up your perspective. Ensure that your argument is structured, formal, and focused. Please avoid addressing counterarguments at this stage.}

\textbf{Review and Discussion by Other Agents:}\\
\textit{You are tasked with reviewing the initial argument presented by Agent 1 on the topic `\textbf{[Insert topic here]}'. Please provide constructive feedback on the argument, discussing its strengths and weaknesses. You may agree, disagree, or partially support the view, but you must justify your stance with reasoned arguments, counterexamples, or further supporting evidence. Be respectful and formal in your review. The following is the `\textbf{[Initial Statement]}}'.

\textbf{Revision:}\\
\textit{Now that you have received feedback from Agents 2–6 on your initial argument (The following is the `\textbf{[Reviews]}'.), you are tasked with revising your viewpoint. Please take into account the points raised by the reviewers, addressing critiques or incorporating any valuable insights. Your revised argument should be more refined, considering both the strengths and weaknesses highlighted in the reviews. Provide a final statement on your position.}

\textbf{Required Answer Format:}\\
\textit{Specify any format requirements for the answer, such as single-word, multiple choice, detailed explanation, list, etc.}"

\begin{center}\textbf{--- Prompt 2 ---}\end{center}

``\textbf{Task: Generate Answer for Specific Tasks}

\textbf{NER:}\\
\textit{You are tasked with performing Named Entity Recognition (NER). Given the following sentence/passage, identify and classify all the named entities. Entities should be categorized as Person, Organization, Location, Date, or Miscellaneous. Sentence/Passage: \textbf{[Insert sentence or passage]}. Your Response: Provide a list of entities along with their respective categories.}\\
\texttt{EXAMPLE:
\textbf{Example Input:}
Sentence: “Barack Obama was born on August 4, 1961, in Honolulu, Hawaii.”
\textbf{Expected Output:}
Barack Obama: Person
August 4, 1961: Date
Honolulu: Location
Hawaii: Location}

\textbf{RE:}\\
\textit{You are tasked with performing Relation Extraction (RE). Given the sentence or passage, identify the entities and their relationship to one another. Specify the type of relationship (e.g., works for, located in, born in). Sentence/Passage: \textbf{[Insert sentence or passage]}. Your Response: Identify the entities and describe the relationship between them.}\\
\texttt{EXAMPLE:
\textbf{Example Input:}
Sentence: “Steve Jobs founded Apple in 1976 in Cupertino, California.”
\textbf{Expected Output:}
Steve Jobs [Person] – Founded – Apple [Organization]
Apple [Organization] – Located in – Cupertino, California [Location]}

\textbf{TC:}\\
\textit{You are tasked with performing Text Classification (TC). Given the sentence or passage, classify it into one of the following categories: [Insert categories, e.g., positive/negative, news/sports/entertainment]. Sentence/Passage: \textbf{[Insert sentence or passage]}. Your Response: Provide the category that best fits the content of the text.}\\
\texttt{EXAMPLE:
\textbf{Example Input:}
Sentence: “The weather today is absolutely beautiful and sunny.”
Categories: Positive, Negative
\textbf{Expected Output:}
Positive}

\textbf{QA:}\\
\textit{You are tasked with performing Question Answering (QA). Given the passage and the question, provide a concise and accurate answer. Passage: \textbf{[Insert passage]}. Question: \textbf{[Insert question]}. Your Response: Provide the answer based on the passage.}\\
\texttt{EXAMPLE:
\textbf{Example Input:}
Passage: “Marie Curie was a physicist and chemist who conducted pioneering research on radioactivity. She was the first woman to win a Nobel Prize.”
Question: “Who was the first woman to win a Nobel Prize?”
\textbf{Expected Output:}
Marie Curie}"

\begin{center}\textbf{--- Prompt 3 ---}\end{center}

``\textbf{Task: Distinguish Augmented Data and Original Data}

\textit{You are tasked with distinguishing between augmented data and original data based on four key aspects: word definition, word-word similarity, syntax correctness, and word examples. For each aspect, follow the definitions provided and apply the evaluation rule to identify whether the data has been augmented or is original.}\\
\textbf{Original Data: `[Insert here]'}\\
\textbf{Augmented Data: `[Inser here]'}\\
\textbf{Word Definition}\\
\textit{\textbf{Definition:} A word definition refers to the precise meaning or set of meanings attributed to a word. In original data, definitions typically align with conventional or dictionary meanings, while in augmented data, meanings may be modified slightly or appear less conventional due to transformations.\\
\textbf{Evaluation Rule:} Compare the meaning of words in the data with their standard definitions. If words seem to deviate from their conventional meanings, or if less common or rephrased meanings are present, it could indicate augmented data.}\\
\textbf{Word-Word Similarity}\\
\textit{\textbf{Definition:} Word-word similarity measures how closely related two words are in meaning, either semantically or contextually. In original data, words tend to reflect naturally occurring relationships. Augmented data may introduce words with forced or unusual similarities due to transformations such as synonym replacement.\\
\textbf{Evaluation Rule:} Assess the semantic relationship between words in context. If word pairs exhibit unnatural or lower similarity compared to typical usage, or if unusual word choices are used to maintain similarity, this may indicate augmented data.}\\
\textbf{Syntax Correctness}\\
\textit{\textbf{Definition: }Syntax correctness refers to the adherence to the grammatical structure of sentences. Original data follows standard syntactical rules, whereas augmented data may introduce slight errors or unusual patterns due to transformations that alter sentence structure.\\
\textbf{Evaluation Rule: }Analyze sentence structure for grammatical accuracy. If there are noticeable shifts in word order, incorrect verb forms, or awkward phrasing that breaks typical syntactical patterns, this suggests augmented data.}\\
\textbf{Word Examples}\\
\textit{\textbf{Definition:} Word examples are common instances of how a word is used in context. Original data will present examples that are conventional and contextually appropriate, while augmented data may use less common or slightly mismatched examples due to changes in phrasing or context.\\
\textbf{Evaluation Rule:} Evaluate whether the words in the data are used in typical and contextually correct examples. If word usage appears slightly out of place, with less common or unconventional examples, this could indicate augmented data.}"

\bibliography{custom}
\bibliographystyle{IEEEtran}

\end{document}